\pdfoutput=1
\documentclass[11pt]{article}
\usepackage[table]{xcolor}% http://ctan.org/pkg/xcolor

\usepackage{acl}

\usepackage{times}
\usepackage{latexsym}

\usepackage[T1]{fontenc}
\usepackage[utf8]{inputenc}

\usepackage{microtype}

\usepackage{url}
\usepackage{easyReview}

\usepackage{cleveref}
\usepackage{multirow}
\usepackage{tabularx}

\usepackage{graphicx}

\usepackage{subfig}% http://ctan.org/pkg/subfig
\usepackage{booktabs}

\usepackage{caption}
\usepackage{tikz}

\definecolor{cadetgrey}{rgb}{0.57, 0.64, 0.69}
\definecolor{greygrey}{rgb}{0.169,0.169,0.169}

\title{\textit{Idioms, Probing and Dangerous Things}: \\ 
Towards Structural Probing for Idiomaticity in Vector Space}
\author{Filip Klubička, Vasudevan Nedumpozhimana, John D. Kelleher \\
  ADAPT Centre, Technological University Dublin, Ireland \\
  \texttt{filip.klubicka@adaptcentre.ie} \\ \texttt{\{vasudevan.nedumpozhimana,john.d.kelleher\}@TUDublin.ie} 
  }

\begin{document}
\maketitle
\begin{abstract}

The goal of this paper is to learn more about how idiomatic information is structurally encoded in embeddings, using a structural probing method. We repurpose an existing English verbal multi-word expression (MWE) dataset to suit the probing framework and perform a comparative probing study of static (GloVe) and contextual (BERT) embeddings. Our experiments indicate that both encode some idiomatic information to varying degrees, but yield conflicting evidence as to whether idiomaticity is encoded in the vector norm, leaving this an open question. We also identify some limitations of the used dataset and highlight important directions for future work in improving its suitability for a probing analysis. % "We find conflicting evidence as to the question of the norm's importance in encoding idiomaticity and suspect the usd dataset is not well-suited to our framework, requiring significant reworks."
\end{abstract}

\section{Introduction}

In recent years the NLP community has become somewhat enamoured by research on probing vector embeddings \cite{ettinger-etal-2016-probing,shi-etal-2016-string,veldhoen2016diagnostic,adi2017fine} and justifiably so, as the method allows researchers to explore linguistic aspects of text encodings and has broad application potential. To date, however, the majority of impactful probing work focuses on analysing syntactic properties encoded in language representations, and the rich and complex field of semantics is comparably underrepresented \cite{belinkov-glass-2019-analysis,rogers2020primer}. One semantic problem that has received relatively little attention is the question of how models encode idiomatic meaning. 

Laterally, our recently-developed extension of the probing method called \textit{probing with noise} \citep{klubicka-kelleher-2022-probing} allows for structural insights into embeddings, highlighting the role of the vector norm in encoding linguistic information and showing that the norm of various embeddings can contain information on various surface-level, syntactic and contextual linguistic properties, as well as taxonomic ones \cite{klubicka-2023-taxonomic}. 

We hypothesise that probing idiomatic usage is a relevant task for studying the role of the norm: given there is some agreement that idiomatic phrases are at least partially defined by how strongly they are linked to the cohesive structure of the immediate discourse \citep{Sag:2002,fazly:2009,sporleder:2009,Feldman:2013,king-cook-2017-supervised}, our intuition is that an idiomatic usage task should behave similarly to contextual incongruity tasks such as bigram shift and semantic odd-man-out \cite{conneau-etal-2018-cram}, which have been shown to be at least partially stored in BERT's vector norm \cite{klubicka-kelleher-2022-probing}. For example, the idiomatic usage of a phrase such as \textit{spill the beans} should have a similarly confounding effect on the sentence's word co-occurrence statistics as a semantic odd-man-out. This reasoning aligns with the findings of \citet{nedumpozhimana-kelleher-2021-finding} who find that BERT can distinguish between sentence disruptions caused by missing words and the incongruity caused by idiomatic usage. Based on this, we are inclined to view an idiomatic usage task as a contextual incongruity task, and would expect to find some information stored in the norm. 

To study this we repurpose an existing idiom token identification dataset into a probing task dataset and run it through our \textit{probing with noise} pipeline, using both static GloVe and contextual BERT embeddings. Interestingly, while our experiments show that both GloVe and BERT generally do encode some idiomaticity information, the norm's role in this encoding is inconclusive, and further analysis points to some surprising irregularities in the behaviour of the models, which we trace back to a number of limitations in the dataset.

\section{Related Work}
\label{s:rw}

Probing in NLP is defined by \citet{conneau-etal-2018-cram} as a classification problem that predicts linguistic properties using dense embeddings as training data. The idea is to train a classifier over embeddings produced by some pretrained model, and assess the embedding model’s knowledge encoding via the probe’s performance. The framework rests on the assumption that the probe's success at a given task indicates that the encoder is storing information on the pertinent linguistic properties.

Given that embeddings are vectors positioned in a shared multidimensional vector space, we are interested in the structural properties of the linguistic information that they encode. Vectors are geometrically defined by two aspects: having both a \textbf{direction} and \textbf{magnitude}. Direction is the position in the space that the vector points towards (expressed by its dimension values), while magnitude is a vector's length, defined as its distance from the origin (expressed by the vector norm). Information contained in a vector is commonly understood to be encoded in the dimension values, however we have shown that it is also possible for the vector magnitude---the norm---to carry information as well \cite{klubicka-kelleher-2022-probing}.

This is an important consideration for embedding research as it has been shown that normalising vectors removes information encoded in the norm \citep{goldberg2017neural,klubicka-kelleher-2022-probing}. A key step in calculating a cosine similarity measure, which is commonly used as a proxy for word similarity, is to normalise the vectors being compared. This has the side effect of nullifying any distinguishing properties the norms might have %, reducing the comparison to one of directions (i.e. dimensions), 
and any linguistic information encoded in the norm will be lost when making the comparison, which is an undesirable outcome if one wished to consider it in the comparison. %Knowing whether an encoder stores some type of information in the vector norm can inform approaches and methods involving vector research. 
We are thus interested in exploring how idiomaticity is encoded in vector space and whether any of it can be found in the norm.

The term Multi-Word Expression (MWE) frequently encompasses a wide variety of phenomena such as idioms, compound nouns, verb particle constructions, etc. The precise definition sometimes differs depending on the community of interest \citep{constant-etal-2017-survey}, and in this paper we use the terms \textit{MWE}, \textit{idiom} and \textit{idiomatic phrase} somewhat liberally to mean any construction with idiomatic or idiosyncratic properties. This is sufficient for our interest in probing for a general notion of idiomaticity, the difference between idiomatic and literal usage of MWEs and studying how this distinction is encoded by embedding models. 

Notably, as probing is a relatively recent framework and idioms are still a difficult phenomenon to model, not much work has been done in this space. Some inspiration can be found in the idiom token identification literature, closely related to word-sense disambiguation, where the goal is to build models that can discriminate idiomatic from literal usage \citep{Hashimoto:2008,Hashimoto:2009,fazly:2009,Li:2010:Cues,Li:2010:Gauss,peng-etal-2014-classifying,salton-etal-2017-idiom,Peng:2017,king-cook-2018-leveraging,shwartz-dagan-2019-still,hashempour-villavicencio-2020-leveraging}. While they do not overtly apply probing in their work, \citet{salton-etal-2016-idiom} were the first to use an idiom token identification pipeline that is comparable to a typical probing framework, where sentence embeddings are used as input to a binary classifier that predicts whether the sentence contains a literal or figurative use of a MWE, indicating that an idiom probing task can be successful.

We have built upon this notion and performed sentence-level probing for idiomaticity in BERT \citep{nedumpozhimana-etal-2022-shapley}. We employed the game theory concept of Shapley Values \citep{shapley1953} to rank the usefulness of individual idiomatic expressions for model training, in an effort to identify the types of signal that BERT captures when modelling idiomaticity. This approach has revealed that providing training data that maximises coverage across topics is the most useful form of topic information, and our findings indicate that there is no one dominant property that makes an expression useful, but rather fixedness and topic features are combined contributing factors. This current paper presents a successor study, as we now look for structural traces of idiomaticity at the sentence level. However, recently there have also been some interesting word-level probing studies.

\citet{nedumpozhimana-kelleher-2021-finding} perform word-level probing experiments on BERT, where they combine probing with input masking to analyse the source of idiomatic information in a sentence, and what form it takes. Results indicate that BERT’s idiomatic key is primarily found within an idiomatic expression, but also draws on information from the surrounding context. Meanwhile, \citet{garcia-etal-2021-probing} use probing to assess if some of the expected linguistic properties of idiomatic noun compounds and their dependence on context and sensitivity to lexical choice can be extracted from contextual embeddings. They conclude that idiomaticity is not yet accurately represented by contextual models: while they might be able to detect idiomatic usage, they may not detect that idiomatic noun compounds have a lower degree of substitutability of their individual components. % when compared to more compositional phrases. % Interestingly, this finding seems to contrast with the findings of \citet{nedumpozhimana-kelleher-2021-finding}, who show evidence that BERT does encode some representation of idiomaticity. We suspect this discrepancy could be due to the use of different datasets and probing altogether different types of idiomatic expressions: one set of experiments works on noun-noun compounds, while the other works on verb-noun phrases.

When it comes to idiomatic probing benchmarks, the Noun Compound Senses Dataset \cite{garcia-etal-2021-probing} is the only curated idiomaticity probing dataset. Other idiom probing work \citep{salton-etal-2016-idiom,nedumpozhimana-kelleher-2021-finding,nedumpozhimana-etal-2022-shapley} relies on existing MWE and idiom datasets, specifically the VNC-tokens dataset \citep{Cook:2008}. Other MWE resources for English include the PARSEME working group's \citep{savary-etal-2017-parseme,ramisch-etal-2018-edition} VMWE dataset, \citep{walsh-etal-2018-constructing}, the STREUSLE corpus \citep{schneider-smith-2015-corpus} and a verbal MWE dataset by \citet{kato-etal-2018-construction}. However, these are annotated at the word-level, employ a fine-grained taxonomy of labels and only annotate idiomatic usage of MWEs, making it impossible to train models that can differentiate between literal and idiomatic usage. As such, while meticulously crafted and, as we argue in \S\ref{s:limitations}, of much higher quality than what we use in our work, they are not suited for the type of sentence-level analysis of idiomaticity we are interested in. There are recent datasets that are better suited for this: MAGPIE \cite{haagsma-etal-2020-magpie} and the SemEval-2022 Task 2 dataset \cite{tayyar-madabushi-etal-2022-semeval}. Unfortunately we only became aware of the former during the review process, while the latter was not yet freely available at the conception of this research. Instead, to stay consistent with the recent wave of idiom probing work, we repurpose the the VNC-tokens dataset \citep{Cook:2008} to suit our structural probing needs, as presented in \S\ref{s:dataset}.

\section{Probing Dataset Construction}
\label{s:dataset}

%\citet{conneau-etal-2018-cram} state that a probing task needs to ask a simple, non-ambiguous question, in order to minimise interpretability problems and confounding factors. 

Our \textit{Idiomatic Usage} (IU) task is based on the VNC-Tokens dataset \citep{Cook:2008}, which is a collection of English sentences containing MWEs called Verb-Noun Combinations (VNC), which can be used idiomatically or literally. This includes expressions such as \textit{hit road}, \textit{blow whistle}, \textit{make scene} and \textit{make mark}. The VNC-tokens dataset contains a total of 2,984 sentences with 56 different expressions, with each sentence containing one expression. Each sentence in the dataset is labelled as \textit{Idiomatic}, \textit{Literal}, or \textit{Unknown}. However, the related literature only makes use of a subset of the full dataset. For consistency and comparability with related work \citep{peng-etal-2014-classifying,salton-etal-2016-idiom,nedumpozhimana-kelleher-2021-finding} we apply the same filtering heuristics %: we remove all sentences labelled as \textit{Unknown} from the dataset before running experiments. Furthermore, out of the 56 different idiomatic expressions, only 28 are considered to have a balanced ratio of idiomatic and literal usage in the example sentences, while the remaining 28 idiomatic expression have a skewed ratio. As such, the latter samples are not considered suitable for experiments in the literature. We thus use the subset of 28 VNCs considered to have a balanced ratio, where roughly 60\% of instances across the dataset are labelled as idiomatic. After these data preparation steps, 
so the subset used in our experiments contains 1,205 sentences, of which 749 are labelled as \textit{Idiomatic} and 456 are labeled as \textit{Literal}, allowing for straightforward binary classification. A breakdown of each expression in the dataset is displayed in Table \ref{tab:dataset} in Appendix \ref{sec:appendix_a}. 

\subsection{Choosing the right train and test split}
\label{ss:vnc_traintest}

In establishing a train and test split we aimed to avoid lexical memorisation \citep{levy-etal-2015-supervised,santus-etal-2016-nine,shwartz-etal-2017-hypernyms}, as our goal is for the probe to only learn a general, abstract, notion of idiomaticity unrelated to any particular idiomatic phrase, so the train and test sets need to be carefully curated. We tackle this on two fronts:

\textbf{(a)} 
The probe needs to be tested on a subset of VNCs that it has not seen in training. Having it predict the usage status of only unfamiliar idiomatic phrases forces the model to fall back on its general knowledge of what makes an idiomatic phrase, rather than a memory of any specific VNC.

\textbf{(b)} 
When training, we also need to ensure that the model attends to general properties of idiomaticity, rather than phrase- or token-specific ones. The surface form of a VNC likely has significant informational value to either the encoder or the probe, so specific VNC constituents might be interpreted as some sort of signal. Upon inspection of the candidate phrases we have found that many of the 28 VNCs in the dataset share the same verb constituent, as shown in Table \ref{tab:verbs}. In fact, the dataset contains only 7 VNCs that contain “unique” verb constituents: \textit{hold fire}, \textit{have word}, \textit{take heart}, \textit{kick heel}, \textit{see star}, \textit{cut figure}, \textit{find foot}.

\begin{table}
\centering
\begin{tabular}{l|l}
verb & noun \\
\hline
\textbf{make} & face, pile, hay, scene, mark, hit \\
\textbf{pull} & leg, weight, plug, punch \\
\textbf{blow} & whistle, top, trumpet \\
\textbf{hit} & wall, roof, road \\
\textbf{get} & wind, sack, nod \\
\textbf{lose} & head, thread \\
\end{tabular}
\caption{Groups of VNCs based on verb constituent overlap.}
\label{tab:verbs}
\end{table}

% This verbal overlap might be interpreted as a signal---were we to include different VNCs containing the same verb in both the train and test set, the probe might recognise the verb and yet again rely on its similarity with what it has encountered during training to make a prediction.

We attempt to mitigate this by populating the train set exclusively with phrases with overlapping verbs, while placing the phrases with unique verbs in the test set. Thus the importance of individual verbs is reduced as they appear with different nouns. Coincidentally, satisfying condition (b) also satisfies condition (a), so no additional filtering is needed: VNCs from the test set do not appear in the train set, and the usage of verbs in the train set is diverse with different VNCs having the same verb constituent. As such, our test set includes 7 VNCs, while the remaining 21 are used in training. Table \ref{tab:traintest} in Appendix \ref{sec:appendix_a} shows the final train and test split used in our experiments. 

Additionally, to confirm that the chosen train and test split is a viable way to tease out idiomaticity, we also run a parallel set of experiments using a form of bootstrapping where we resample the train and test split multiple times by randomly choosing 7 VNCs to be used in the test set, and using the remaining 21 phrases for training. This violates the above-established principle (b) as verbal constituents might be mixed between train and test sets, but still conforms to principle (a), as the model will always be tested on a set of 7 phrases that were not seen during training. Additionally, as we are not fixing the number of samples in the train and test sets, but rather the number of idiomatic phrases (with a varying number of sentences containing each phrase), there will also be slight differences in the ratio of the train and test sample sizes between different runs. However, we find that when the multitude of runs are averaged the true effect comes to the fore---the bootstrapped results mirror the results of the fixed setting, confirming the chosen split. %\footnote{In fact, a Pearson correlation analysis between the train and test sample sizes and the obtained evaluation scores yields a coefficient no higher or lower than +/- 0.026, showing no correlation.}
For transparency and completeness, in Section \ref{s:results} we report results for both setups: Idiomatic Usage Fixed data split (IU\textsubscript{F}) and Idiomatic Usage Resampled data split (IU\textsubscript{R}).

\section{Experimental Setup}
\label{s:models}

\subsection{Chosen Embeddings}

Given the prominence of contextual encoders such as BERT \citep{devlin-etal-2019-bert} and its derivatives, as well as their ability to model in-context meaning and incongruity, they are an obvious choice for our analysis. However, rather than compare different contextual encoders, we prefer to draw a contrastive comparison with a static encoder such as GloVe \cite{Pennington:2014}, which is based on a word to word co-occurrence matrix, as this comparison can provide more varied insight.

Given that our idiomatic usage dataset is framed as a classification task at the sentence level, our experiments require sentence representations. We use pretrained versions of BERT and GloVe to generate embeddings for each sentence. The BERT model generates 12 layers of embedding vectors with each layer containing a separate 768-dimensional embedding for each word, so we average the word embeddings in BERT's final layer, resulting in a 768-dimensional sentence embedding. We take the same mean pooling approach with GloVe, which yields a 300-dimensional sentence embedding for each sentence. While BERT uses sub-word tokens to get around out of vocabulary tokens, in the rare instance of encountering an OOV with GloVe, we generate a random word embedding in its stead.

\subsection{Probing with Noise}

The method is described in detail in \citet{klubicka-kelleher-2022-probing}\footnote{Code available here: \url{https://github.com/GreenParachute/probing-with-noise}}: in essence it applies targeted noise functions to embeddings that have an ablational effect and remove information encoded either in the norm or dimensions of a vector.

We remove information from the norm (abl.N) by sampling random norm values and scaling the vector dimensions to the new norm. Specifically, we sample the L2 norms uniformly from a range between the minimum and maximum L2 norm values of the respective embeddings in our dataset.\footnote{GloVe: [2.2634,4.2526]\\
BERT: [7.4844,11.1366]}

To ablate information encoded in the dimensions (abl.D), we randomly sample dimension values and then scale them to match the original norm of the vector. Specifically, we sample dimension values uniformly from a range between the minimum and maximum dimension values of the respective embeddings in our dataset.\footnote{GloVe: [-1.7866, 2.8668]\\
BERT: [-5.0826, 1.5604]} We expect this to fully remove all interpretable information encoded in the dimension values, making the norm the only information container available to the probe. 

Applying both noise functions to the same vector (abl.D+N) should remove any information encoded in it, meaning the probe has no signal to learn from, a scenario equal to training on random vectors. 

Even when an embedding encodes no information, our train set contains class imbalance and the probe can learn the distribution of classes. To account for this, as well as the possibility of a powerful probe detecting an empty signal \citep{zhang-bowman-2018-language}, we establish informative random baselines against which we compare the probe's performance. We employ two such baselines: (a) we assert a random prediction (\textit{rand.pred}) onto the test set, negating any information that a classifier could have learned, class distributions included; and (b) we train the probe on randomly generated vectors (\textit{rand.vec}), establishing a baseline with access only to class distributions. 

Finally, to address the degrees of randomness in the method, we train and evaluate each model 50 times and report the average score of all the runs, essentially bootstrapping over the random seeds \citep{wendlandt-etal-2018-factors}. Additionally, we calculate a confidence interval (CI) to ensure that the reported averages were not obtained by chance, and report it alongside the results to indicate statistical significance when comparing averages.

\subsection{Probing Classifier and Evaluation Metric}

In our experiments the sentence embeddings are used as input to a Multi-Layered Perceptron (MLP) classifier, which labels them as idiomatic (1) or literal (0). We evaluate the performance of the probe using the micro-average AUC-ROC score,\footnote{\url{https://scikit-learn.org/stable/modules/generated/sklearn.metrics.roc_auc_score.html}} the most appropriate evaluation metric for a dataset with unbalanced labels, as it reflects the classifier's performance on both positive and negative classes. Regarding implementation and parameter details, we used the bert-base-uncased BERT model from the \textit{pytorch\_pretrained\_bert} library\footnote{\url{https://pypi.org/project/pytorch-pretrained-bert/}} \citep{PyTorch}, a pre-trained GloVe model\footnote{The larger common crawl vectors: \url{https://nlp.stanford.edu/projects/glove/}} and for the MLP probe we used the scikit-learn MLP implementation \citep{scikit-learn} using the default parameters.\footnote{activation='relu', solver='adam', max\_iter=200, hidden\_layer\_sizes=100, learning\_rate\_init=0.001, batch\_size=min(200,n\_samples), early\_stopping=False, weight init. $W \sim \mathcal{N}\left(0, \sqrt{6/(fan_{in}+fan_{out})}\right)$ (scikit relu default). See:  \url{https://scikit-learn.org/stable/modules/generated/sklearn.neural_network.MLPClassifier.html} }

\section{Experimental Results}
\label{s:results}

Experimental evaluation results for GloVe and BERT on the idiomatic usage (IU) probing task are presented in Tables \ref{tab:vnc-results-glove} and \ref{tab:vnc-results-bert}. The tables include results for both the setting where the VNC's in the hold-out test set are fixed (IU\textsubscript{F}) and the setting where they are resampled each time (IU\textsubscript{R}), though this is essentially the same probing task. Note that cells shaded light grey belong to the same distribution as random baselines, as there is no statistically significant difference between the different scores; cells shaded dark grey belong to the same distribution as the vanilla baseline; and cells that are not shaded contain a significantly different score than both the random and vanilla baselines, indicating that they belong to different distributions.

The results interpretation here is quite straightforward. As the unablated, vanilla baseline significantly outperforms random baselines in both models, this indicates that both GloVe and BERT encode a non-zero amount of idiomatic usage information, which aligns with previous findings.

\begin{table}
\centering
\scalebox{0.95}{
\centering
\begin{tabular}{|l|c|c|c|c|}
\hline
\multicolumn{5}{|c|}{\textbf{GloVe}} \\
\hline
Model & \multicolumn{2}{m{4em}|}{\textbf{IU\textsubscript{F}}} & \multicolumn{2}{m{4em}|}{\textbf{IU\textsubscript{R}}} \\
 & auc & \textpm CI & auc & \textpm CI \\
\hline
rand. pred. & \cellcolor{cadetgrey!25} .4994 & \cellcolor{cadetgrey!25} .0015 & \cellcolor{cadetgrey!25} .4998 & \cellcolor{cadetgrey!25} .0013 \\
rand. vec. & \cellcolor{cadetgrey!25} .4997 & \cellcolor{cadetgrey!25} .0015 & \cellcolor{cadetgrey!25} .5 & \cellcolor{cadetgrey!25} .0013 \\
\hline
vanilla & \cellcolor{greygrey!25} .7485 & \cellcolor{greygrey!25} .0003 & \cellcolor{greygrey!25} .7717 & \cellcolor{greygrey!25} .0022 \\
\hline
abl. N & .7445 & .0006 & \cellcolor{greygrey!25} .7687 & \cellcolor{greygrey!25} .0021 \\
abl. D & \cellcolor{cadetgrey!25} .5012 & \cellcolor{cadetgrey!25} .0018 & \cellcolor{cadetgrey!25} .4993 & \cellcolor{cadetgrey!25} .0015 \\
abl. D+N & \cellcolor{cadetgrey!25} .4991 & \cellcolor{cadetgrey!25} .0018 & \cellcolor{cadetgrey!25} .5005 & \cellcolor{cadetgrey!25} .0015 \\
\hline
\end{tabular}}
\caption{Probing results on GloVe models and baselines, both with fixed (F) and resampled (R) test set. Reporting average AUC-ROC scores and confidence intervals (CI) of the average of all training runs.}
\label{tab:vnc-results-glove}
\end{table}

\begin{table}
\centering
\scalebox{0.95}{
\centering
\begin{tabular}{|l|c|c|c|c|}
\hline
\multicolumn{5}{|c|}{\textbf{BERT}} \\
\hline
Model & \multicolumn{2}{m{4em}|}{\textbf{IU\textsubscript{F}}} & \multicolumn{2}{m{4em}|}{\textbf{IU\textsubscript{R}}} \\
 & auc & \textpm CI & auc & \textpm CI \\
\hline
rand. pred. & \cellcolor{cadetgrey!25} .4997 & \cellcolor{cadetgrey!25} .0015 & \cellcolor{cadetgrey!25} .4998 & \cellcolor{cadetgrey!25} .0013 \\
rand. vec. & \cellcolor{cadetgrey!25} .4997 & \cellcolor{cadetgrey!25} .0015 & \cellcolor{cadetgrey!25} .5013 & \cellcolor{cadetgrey!25} .0013 \\
\hline
vanilla & \cellcolor{greygrey!25} .8411 & \cellcolor{greygrey!25} .0002 & \cellcolor{greygrey!25} .8524 & \cellcolor{greygrey!25} .0016 \\
\hline
abl. N & \cellcolor{greygrey!25} .8413 & \cellcolor{greygrey!25} .0003 & \cellcolor{greygrey!25} .8532 & \cellcolor{greygrey!25} .0016 \\
abl. D & \cellcolor{cadetgrey!25} .4991 & \cellcolor{cadetgrey!25} .0019 & \cellcolor{cadetgrey!25} .4978 & \cellcolor{cadetgrey!25} .0015 \\
abl. D+N & \cellcolor{cadetgrey!25} .4999 & \cellcolor{cadetgrey!25} .0018 & \cellcolor{cadetgrey!25} .5004 & \cellcolor{cadetgrey!25} .0015 \\
\hline
\end{tabular}}
\caption{Probing results on BERT models and baselines, both with fixed (F) and resampled (R) test set. Reporting average AUC-ROC scores and confidence intervals (CI) of the average of all training runs.}
\label{tab:vnc-results-bert}
\end{table}

\textbf{IU\textsubscript{F} vs. IU\textsubscript{R}:} It important to validate our chosen train and test split (see \S\ref{ss:vnc_traintest}) by comparing the respective vanilla performances of IU\textsubscript{F} and IU\textsubscript{R}. Given that our goal is to nudge the probe to model a representation of idiomaticity that is unrelated to any given phrase, we expect that the IU\textsubscript{F} setting should make the task more difficult for the classifier. The results confirm this, showing that in GloVe and BERT vanilla IU\textsubscript{R} significantly outperforms vanilla IU\textsubscript{F}. Evidently, the curated test split makes prediction on the task more challenging and the lower performance of IU\textsubscript{F} indicates that the model is forced to rely on VNC-independent features to make predictions. % In their respective intrinsic evaluations IU\textsubscript{F} and IU\textsubscript{R} exhibit the same behaviour in BERT, while there is only one difference in GloVe, namely that ablating just the norm causes a statistically significant drop in performance in IU\textsubscript{F}, while this is not the case in IU\textsubscript{R}.

\textbf{GloVe vs. BERT:} In terms of differences between encoders, the results show that vanilla BERT significantly outperforms vanilla GloVe in both the IU\textsubscript{F} and IU\textsubscript{R} scenarios. Evidently, BERT is much better at encoding idiomaticity than GloVe. We suspect this is due to two factors: (a) BERT is a contextual encoder and as such is better suited to modelling the local context necessary to accurately represent idiomaticity in the sentence, and (b) it has a much higher dimensionality, meaning it has the potential to devote more representation space to more complex phenomena.

\textbf{Idiomaticity and the norm:} One of the goals of this experiment was to investigate whether the norm encodes any information relevant to the IU task. Our method states this is most clearly determined in the setting with ablated dimension information (abl.D), where above random performance indicates that the information is stored in the unablated norm container \citep{klubicka-kelleher-2022-probing}. Our results here show no conclusive indication that the norm encodes idiomaticity information on this task: in all four scenarios ablating only the dimensions already makes the probe's performance comparable to random, which indicates no information is stored in the norm.\footnote{We do see a hint of this when ablating the norm in GloVe IU\textsubscript{F}, but this is more likely a feature of this particular data split, as the signal is not mirrored in IU\textsubscript{R}. Even if it was, without a signal in the abl.D setting, the abl.N setting is insufficient evidence to infer that the norm encodes information.} 

As stated in the introduction, given the IU task's similarity with contextual incongruity tasks, we would expect to find some signal in the norm. Our result here is somewhat surprising and motivates further questions, prompting us to perform additional post hoc investigations and analyses that should improve our understanding of the results and help shape our overall findings.

\section{Additional Experiments}
\label{s:post-hoc}

\subsection{Norm Correlation Analysis}
\label{ss:norm}

For another perspective on the relationship between vector norms and the IU task information, we run a post hoc analysis on the norm container. We investigate both the norms of our embeddings using a Pearson correlation analysis, which can be considered a linear probing study: we test the correlation between each vector norm (L1 and L2) and the sentence labels (\textit{Idiomatic} and \textit{Literal}\footnote{The Pearson test only works on continuous variables, but it is still possible to calculate with categorical variables if they are binary, by simply converting the categories to 0 and 1.}). The correlation results are presented in Table \ref{tab:correlation} and seem to be somewhat at odds with our experimental results. 

The analysis shows that in both vanilla GloVe and BERT both norms have a weak negative correlation with IU labels. While the correlations are weak, they are not zero---we observe a significant drop in the coefficients upon applying the norm ablation function, which seems to fully remove information from both norms, as the correlation coefficients drop to $\approx$0, indicating that relevant information encoded in the norms has been removed.

This difference between vanilla and abl.N points to some slight correlation between the idiomaticity labels and information encoded in the vanilla norm, yet our probing experiments do not align with this finding. What makes this more unusual is that our IU correlations are comparable to the correlations on parse tree depth (0.1908) or semantic odd-man-out (0.2305) tasks which do produce a signal in the \textit{probing with noise} experiments as previously reported \citet{klubicka-kelleher-2022-probing}.

\begin{table}[tb]
\centering
\scalebox{0.80}{
\centering
\begin{tabular}{|l|l|c|c|c|c|}
\hline
Task & Vectors & \multicolumn{2}{m{5.2em}|}{\textbf{GloVe}} & \multicolumn{2}{m{5.2em}|}{\textbf{BERT}} \\
 & & L1 & L2 & L1 & L2 \\
\hline
& vanilla & -0.2231 & -0.1786 & -0.1490 & -0.1756 \\
IU & abl. N & -0.0074 & 0.0276 & -0.0397 & -0.0167 \\

\hline
\end{tabular}}
\caption{Pearson correlation coefficients between class labels and L1 and L2 norms for vanilla vectors and vectors with ablated norms. For this analysis the Idiomatic label was mapped to 1 and the Literal label to 0.}
\label{tab:correlation}
\end{table}

It is possible that the correlation is just on the verge of being too weak to be detectable by the method. On the other hand, this could be a sign that other factors are at play---we suspect that the misalignment between the probing and correlation results hints at the imbalanced nature of the IU dataset and its limitations. We run an additional experiment to search for more evidence.

As an aside, it is worth noting that if we were to take the correlation results at face value, they do provide some interesting insight into how idiomatic usage is encoded in vector space. Specifically, a non-zero negative correlation coefficient means that sentences containing idiomatic usage are positioned closer to the origin relative to sentences that contain literal usage. In other words, both GloVe and BERT vectors of sentences containing idiomatic usage are slightly shorter, which is an intriguing structural finding.

\subsection{Dimension Deletion}\label{ss:deletion_all}

We run supplementary experiments to investigate the role of the dimension container as the sole carrier of IU information. To do this we perform a dimension deletion experiment. Partially inspired by the work of \citet{torroba-hennigen-etal-2020-intrinsic} who found that most linguistic properties are reliably encoded by only a handful of dimensions, we attempt to roughly identify the degree of localisation of information in the vector dimensions. In staying consistent with the ablational nature of the method, we simply delete one half of the vector's dimensions and retrain the probe on the truncated vectors, repeating the process for the remaining half.\footnote{Note that deleting dimensions reduces the dimensionality of the vectors and inherently changes the norm of the vectors, which is why it is important to consider this a post hoc analysis: while the ablation functions are used to identify which information container information is encoded in, dimension deletion presupposes that the information is in the dimension container and functions as a test that helps narrow down where in the dimension container the information is encoded.}

The dimension deletion results are included in Tables \ref{tab:del-vnc-results-glove} and \ref{tab:del-vnc-results-bert}. In these tables the row denoted \textit{del.1h} reports the results for deleting the 1\textsuperscript{st} half of an embedding vector, and \textit{del.2h} reports results for deleting the 2\textsuperscript{nd} half. 
Given that all relevant IU information seems to be encoded in vector dimensions, we expect that deleting half of the vector would cause a significant drop in performance when compared to vanilla. We would also expect a drop in evaluation scores regardless of which half of the vector is deleted. However, our results reveal some rather surprising effects.

\begin{table}
\centering
\scalebox{0.95}{
\centering
\begin{tabular}{|l|c|c|c|c|}
\hline
\multicolumn{5}{|c|}{\textbf{GloVe}} \\
\hline
Model & \multicolumn{2}{m{4em}|}{\textbf{IU\textsubscript{F}}} & \multicolumn{2}{m{4em}|}{\textbf{IU\textsubscript{R}}} \\
 & auc & \textpm CI & auc & \textpm CI \\
\hline
rand. pred. & \cellcolor{cadetgrey!25} .4994 & \cellcolor{cadetgrey!25} .0015 & \cellcolor{cadetgrey!25} .4998 & \cellcolor{cadetgrey!25} .0013 \\
rand. vec. & \cellcolor{cadetgrey!25} .4997 & \cellcolor{cadetgrey!25} .0015 & \cellcolor{cadetgrey!25} .5 & \cellcolor{cadetgrey!25} .0013 \\
\hline
vanilla & \cellcolor{greygrey!25} .7485 & \cellcolor{greygrey!25} .0003 & \cellcolor{greygrey!25} .7717 & \cellcolor{greygrey!25} .0022 \\
\hline
del. 1h & \textbf{.7737} & .0005 & .7553 & .0023 \\
del. 2h & .7043 & .0005 & .7545 & .002 \\
\hline
\end{tabular}}
\caption{Probing results on GloVe dimension deletions both with fixed (F) and randomised (R) test set. Reporting average AUC-ROC scores and confidence intervals (CI) of the average of all training runs.} %Cells shaded light grey belong to the same distribution as random baselines, dark grey cells share the vanilla baseline distribution, while scores significantly different from both the random and vanilla baselines are unshaded.
%In the dimension deletion experiments the significantly lower score is marked with an asterisk, while the scores marked in bold show an improvement in performance compared to vanilla baseline.}
\label{tab:del-vnc-results-glove}
\end{table}

%\textbf{GloVe:} 

%Typically, we would expect the indices of informative dimensions to be arbitrary, yet this result seems to indicate that GloVe localises the information it encodes in favour of placing more informative dimensions at the beginning of the vector.
%This points to redundancies within the dimensions themselves, indicating that not many dimensions are needed to encode specific linguistic features. 

While \textit{del.2h} in GloVe causes the expected performance drop, in IU\textsubscript{F} \textit{del.1h} causes a statistically significant \textit{improvement} when compared to the vanilla baseline (marked in bold). We observe quite a large performance spike, though this is not mirrored in the IU\textsubscript{R} scenario. We might dismiss this as just a strange artefact of the particular IU\textsubscript{F} data split, were it not for the fact that we observe the same behaviour in both IU\textsubscript{F} and IU\textsubscript{R} in BERT, where \textit{del.2h} causes a significant performance drop, but \textit{del.1h} causes a significant spike. 

\begin{table}
\centering
\scalebox{0.95}{
\centering
\begin{tabular}{|l|c|c|c|c|}
\hline
\multicolumn{5}{|c|}{\textbf{BERT}} \\
\hline
Model & \multicolumn{2}{m{4em}|}{\textbf{IU\textsubscript{F}}} & \multicolumn{2}{m{4em}|}{\textbf{IU\textsubscript{R}}} \\
 & auc & \textpm CI & auc & \textpm CI \\
\hline
rand. pred. & \cellcolor{cadetgrey!25} .4997 & \cellcolor{cadetgrey!25} .0015 & \cellcolor{cadetgrey!25} .4998 & \cellcolor{cadetgrey!25} .0013 \\
rand. vec. & \cellcolor{cadetgrey!25} .4997 & \cellcolor{cadetgrey!25} .0015 & \cellcolor{cadetgrey!25} .5013 & \cellcolor{cadetgrey!25} .0013 \\
\hline
vanilla & \cellcolor{greygrey!25} .8411 & \cellcolor{greygrey!25} .0002 & \cellcolor{greygrey!25} .8524 & \cellcolor{greygrey!25} .0016 \\
\hline
del. 1h & \textbf{.8668} & .0002 & \textbf{.8576} & .0016 \\
del. 2h & .8137 & .0003 & .8368 & .0016 \\
\hline
\end{tabular}}
\caption{Probing results on BERT dimension deletions both with fixed (F) and randomised (R) test set. Reporting average AUC-ROC scores and confidence intervals (CI) of the average of all training runs.}
\label{tab:del-vnc-results-bert}
\end{table}

It seems that both GloVe and BERT exhibit a certain degree of information localisation, with a preference for storing relevant IU information in the first half of dimensions, to the point where the second half reduces the overall information quality of the vector. In principle this interpretation is consistent with the findings of \citet{torroba-hennigen-etal-2020-intrinsic} and \citet{durrani-etal-2020-analyzing}, who showed that certain linguistic properties are localised in dimensions of contextual embeddings. However, we remain skeptical and wonder whether our findings reflect how these embeddings truly encode idiomaticity, or whether this is property of this particular dataset. We consider this in the following section.

\section{Discussion and Limitations}
\label{s:discussion}

While the correlation coefficients between both GloVe's and BERT's norm and the IU labels are non-zero, our probe does not seem to be able to leverage this information from the norm. In isolation, the correlation coefficient would have led us to believe that there may be some idiomaticity information encoded in the norm. However, this has not been confirmed by the \textit{probing with noise} method, which when used in conjunction with the correlation analysis offers conflicting evidence.

The performance spikes exhibited in the deletion experiments are somewhat baffling, especially given the stark differences between the GloVe and BERT architectures. However, if the IU task were truly analogous to a contextual incongruity task, then arguably vanilla GloVe should be much worse at encoding IU than shown in our results---by design, an averaged GloVe sentence embedding cannot be aware of word order or relationships between words in a specific context and should perform much more poorly on such tasks, making even vanilla GloVe's performance a result that raises more questions than it answers.

One pertinent consideration regards the fact that our experiments were performed at the sentence level. It is possible that there is a crisper signal in the norm of individual word embeddings (as shown on a word-level taxonomic probing task \cite{klubicka-2023-taxonomic}). Averaging word embeddings to obtain sentence representations may have diluted the signal to the point where it is not detectable by the \emph{probing with noise} method. Replicating our experiments at the word-level, or using more direct sentence representation approaches (such as using BERT's CLS token, doc2vec \cite{le2014distributed} or SentenceBERT \cite{reimers-gurevych-2019-sentence}) might produce a more salient result. 

As it stands, the majority of the results we have observed on the IU dataset behave like surprising outliers that are difficult to explain. This can either be due to strong confounding factors at play that we are not aware of or, perhaps more likely, this is evidence of our suspicion that the dataset is not well-suited for this type of analysis. And while we have learned that vanilla BERT is better at the task than GloVe, the question whether idiomaticity can be encoded in their norms remains an open one. 

\subsection{Dataset Limitations}
\label{s:limitations}

While constructing and experimenting with the VNC-tokens dataset we have become aware of some of its shortcomings. Our main concern is that it is two orders of magnitude smaller than more established probing datasets \cite{conneau-etal-2018-cram}. While we addressed this by increasing the number of training runs and resampling the train and test set, its size still limits what the models are able to learn. Unfortunately, in dealing with an intricate phenomenon such as idioms, considerably-sized corpora are few and far between.\footnote{In fatct, all existing MWE resources are within a comparable size range to the VNC-tokens dataset. Even concatenating them would not nearly approach the size of probing datasets for non-semantic tasks.}

The VNC-tokens dataset is also very limited in scope, containing only a single type of verbal MWE, while other datasets include a wider variety of verbal expressions or compounds involving other parts of speech. It is also worth noting that both idiomatic and literal usages of the VMWEs present in the dataset are relatively frequent in English when compared to other more niche idiomatic phrases. This relative frequency is likely also reflected in the pretrained embeddings and could affect a model's ability to model their idiomaticity, raising the question whether relatively rarer phrases might behave differently. Thus the generalisability of our findings to other idiomatic expressions is uncertain. 

Furthermore, at this point the VNC-Tokens dataset is a relatively older benchmark and there are indications that it has not been as meticulously crafted as more recent MWE datasets. For example, the dataset does not control for sentence length, which could be a strong confounding factor, it contains some typographical errors, even some seemingly incorrect IU annotations, as well as literary language which contains OOV tokens for the pretrained GloVe model. It is our impression that cleaning up the dataset, aligning it with the PARSEME annotation guidelines\footnote{\url{https://parsemefr.lis-lab.fr/parseme-st-guidelines/1.1/?page=010_Definitions_and_scope/020_Verbal_multiword_expressions}}, and updating it with additional examples of sentences containing VNCs in order to better balance the idiomaticity labels would greatly improve its overall quality.

Overall, in spite of our best efforts at mitigating confounders and constructing the right data split for our task, we still wonder whether the dataset is simply too small and too imbalanced to truly be useful in our probing scenario. Given all the limitations we have become aware of over the course of our experimentation it is difficult to decide whether our results are inconclusive due to the dataset, the type of idioms studied, perhaps some unknown limitation of the approach, or are simply a true observation. This makes our partially inconclusive and partially surprising findings somewhat difficult to reconcile with previous work. We thus emphasise the importance of expanding this work to a wider category of idiomatic phrases and ideally folding in all the datasets mentioned in \S\ref{s:rw}---applying \textit{probing with noise} to the datasets individually as well as an amalgamation of datasets would provide a more comprehensive analysis of general idiomaticity encoding and could provide more salient insights. It might also be beneficial to consider other dimensions of idiomaticity in the experimentation and analysis, such as evaluating MWEs that are differentiated with respect to whether or not they carry a metaphorical mapping to literal usages, and whether or not they are grammatical or extragrammatical \citep{fillmore1988regularity}.

\section{Conclusion and Future Work}

In this paper we applied the \textit{probing with noise} method to two different types of word representations---static and contextual---generated by two different embedding algorithms---GloVe and BERT---on a repurposed idiomatic usage probing task, with the aim of obtaining structural insights into the role of the norm encoding idiomatic usage information. 

Overall we detect some mixed signals in our findings, which include that \textbf{(a)} generally both GloVe and BERT encode idiomatic usage information, but BERT encodes more \textbf{(b)} the norm of GloVe and BERT carries no idiomaticity information (or at least this is not recoverable by the probe), even though \textbf{(c)} it seems there is a correlation between the norm length and idiomatic usage in a sentence, where sentences containing idiomatic usage are positioned relatively closer to the origin of the vector space. \textbf{(d)} Additionally, it seems both GloVe and BERT prefer to store idiomatic usage information in the first half of their vectors, to the point where the second half is detrimental to the vector's overall encoding of idiomaticity. Finally, \textbf{(e)} we present these findings with the caveat that they only apply to the VNC-Tokens dataset, which requires a bit of a rework in order to be up to the standard required for a probing framework.

As for our initial research question, we asked whether embeddings models such as BERT might see an idiomatic usage task as being of the same category as a contextual incongruity task.\footnote{This hypothesis inspired the title of the paper, referring to \citet{lakoff1987women} and his work on semantic categories.} Given that vanilla BERT strongly outperforms vanilla GloVe on the task, this could lend some credence to the interpretation that contextual awareness and the ability to model incongruity, which GloVe lacks but BERT excels at, is what improves its idiomaticity encoding. However, evidence is inconclusive and whether the vector norm of either model plays a role in encoding idiomatic information in the same way that it supplements the encoding of contextual incongruity information remains an open question, which we are committed to further pursue in future work. This would involve cleaning the VNC-Tokens dataset and combining it with other existing MWE datasets in a systematic exploration of the structural encoding of idiomaticity. 

% Calls for future work

%Given that the question whether idiomaticity can be encoded in the norm remains open, we are quite keen to find an answer. To begin improving the work in this space, we propose starting with updating the VNC-tokens dataset for idiomatic usage, which has proven to be a somewhat underwhelming resource for idiomaticity probing. What is needed is a deep review and cleaning of the existing annotations, aligning the dataset with the PARSEME annotation guidelines and sourcing additional examples of sentences containing idiomatic and literal examples of the VNCs in the dataset, with the aim of improving the balance of idiomaticity labels. If at all possible, it would be wise to also attempt to control for sentence length, as this could be a confounding factor. This line of work would certainly improve the quality of the dataset, which could be released as version 2.0, specially curated to be a probing task dataset.

%Furthermore, we would be very interested in widening the scope of idiomatic expressions that are studied in the probing literature. To this end, we can do two things: (a) create an amalgam of all existing idiom datasets in order to increase training size and apply our method to probe for a very general encoding of idiomaticity, or (b) apply our method to different datasets individually in order to see whether there are any regularities or perhaps differences in the ways different kinds of idiomatic phrases are encoded in vector space.

\section*{Acknowledgements}

This research was conducted with the financial support of Science Foundation Ireland under Grant Agreements No. 13/RC/2106 and 13/RC/2106\_P2 at the ADAPT SFI Research Centre at Technological University Dublin. ADAPT, the SFI Research Centre for AI-Driven Digital Content Technology, is funded by Science Foundation Ireland through the SFI Research Centres Programme, and is co-funded under the European Regional Development Fund.

\bibliography{anthology,custom,mwe2023-f} %references
\bibliographystyle{acl_natbib}

\appendix

\section{Appendix A}
\label{sec:appendix_a}

\subsection{Dataset Statistics}

\begin{table}
\centering
\begin{tabular}{|p{2.3cm}|p{1.4cm}|p{1.7cm}|p{0.7cm}|}
\hline
Expression & \#samples & \#idiomatic & ratio \\
\hline \hline
see star & 61 & 5 & 0.08 \\
hit wall & 63 & 7 & 0.11 \\
pull leg & 51 & 11 & 0.22 \\
hold fire & 23 & 7 & 0.30 \\
make pile & 25 & 8 & 0.32 \\
blow whistle & 78 & 27 & 0.35 \\
make hit & 14 & 5 & 0.36 \\
get wind & 28 & 13 & 0.46 \\
lose head & 40 & 21 & 0.53 \\
make hay & 17 & 9 & 0.53 \\
make scene & 50 & 30 & 0.60 \\
hit roof & 18 & 11 & 0.61 \\
blow trumpet & 29 & 19 & 0.66 \\
make face & 41 & 27 & 0.66 \\
pull plug & 64 & 44 & 0.69 \\
take heart & 81 & 61 & 0.75 \\
hit road & 32 & 25 & 0.78 \\
kick heel & 39 & 31 & 0.79 \\
pull punch & 22 & 18 & 0.82 \\
pull weight & 33 & 27 & 0.82 \\
blow top & 28 & 23 & 0.82 \\
cut figure & 43 & 36 & 0.84 \\
make mark & 85 & 72 & 0.85 \\
get sack & 50 & 43 & 0.86 \\
have word & 91 & 80 & 0.88 \\
get nod & 26 & 23 & 0.88 \\
lose thread & 20 & 18 & 0.90 \\
find foot & 53 & 48 & 0.91 \\
\hline
TOTAL: & 1205 & 749 & 0.62 \\
\hline
\end{tabular}
\caption{VNCs ordered by \% of idiomatic usage: number of samples (\#samples), number of idiomatic uses (\#idiomatic) \% of idiomatic usage (ratio).
}
\label{tab:dataset}
\end{table}

In Table \ref{tab:dataset} the VNC expressions are listed by increasing order of percentage of idiomatic usage: \textit{see star} is the expression with the lowest percentage of idiomatic usage (8.20\%) and \textit{find foot} is the expression with the highest percentage of idiomatic usage (90.57\%). The overall percentage of idiomatic instances (regardless of the expression) is 62\%.

\begin{table}[tb]
\centering
\scalebox{0.80}{
\centering
\begin{tabular}{|l|c|c||l|c|c|}
\hline
 & \multicolumn{2}{c||}{\textbf{Train set}} &  & \multicolumn{2}{c|}{\textbf{Test set}}  \\
VNC & Total & Idiomatic & VNC & Total & Idiomatic \\
\hline
\hline
blow top & 28 & 23 &  &  &  \\
blow trumpet & 29 & 19 &  &  & \\
blow whistle & 78 & 27 &  &  &  \\
get sack & 50 & 43 &  &  &  \\
get nod & 26 & 23 &  &  &  \\
get wind & 28 & 13 &  &  &  \\
hit road & 32 & 25 &  &  & \\
hit roof & 18 & 11 & cut figure & 43 & 36 \\
hit wall & 63 & 7 & find foot & 53 & 48  \\
lose head & 40 & 21 & have word & 91 & 80 \\
lose thread & 20 & 18 & hold fire & 23 & 7 \\
make face & 41 & 27 & kick heel & 39 & 31 \\
make hay & 17 & 9 & see star & 61 & 5 \\
make hit & 14 & 5 & take heart & 81 & 61 \\
make mark & 85 & 72 &  &  &  \\
make pile & 25 & 8 &  &  &  \\
make scene & 50 & 30 &  &  &  \\
pull leg & 51 & 11 &  &  &  \\
pull plug & 64 & 44 &  &  &  \\
pull punch & 22 & 18 &  &  &  \\
pull weight & 33 & 27 &  &  &  \\
\hline
Total: & 814 & 481 &  & 391 & 268 \\
Ratio: &  & 0.5909 &  &  & 0.6854 \\
\hline
\end{tabular}}
\caption{A breakdown of VNCs and idiomatic instances in the train and test split.}
\label{tab:traintest}
\end{table}

Table \ref{tab:traintest} displays the final train and test split we used in our experiments, as well as a breakdown of specific expressions and their labels in both sets, sorted according to the verbal constituent. While this split is not focused on the ratio of training instances, but rather subsets of training instances containing the same VNC, this does mirror the 25\%/75\% data split employed by \cite{salton-etal-2016-idiom}. Though the 68\% ratio of idiomatic phrases in the test set is somewhat higher than maintained in previous work ($\approx$62\%), we expect the specific choices of VNCs will have a positive effect overall in priming the classifier to use its knowledge of idiomaticity to make predictions.

\end{document}